\documentclass{article}

\usepackage{arxiv}

\usepackage[utf8]{inputenc} 
\usepackage[T1]{fontenc}    
\usepackage{hyperref}       
\usepackage{url}            
\usepackage{booktabs}       
\usepackage{amsfonts}       
\usepackage{nicefrac}       
\usepackage{microtype}      
\usepackage{lipsum}		
\usepackage{graphicx}
\usepackage{natbib}
\usepackage{doi}

\usepackage{csvsimple}
\usepackage{caption}
\usepackage{subcaption}
\usepackage{algorithm}
\usepackage{algpseudocode}
\usepackage{amsmath}
\usepackage{tabularx}
\usepackage[table]{xcolor}
\usepackage{hyperref}
\usepackage{makecell}

\title{Run LoRA Run: Faster and Lighter LoRA Implementations}

\author{ \href{https://orcid.org/0000-0000-0000-0000}{\includegraphics[scale=0.06]{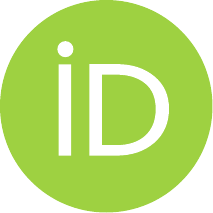}\hspace{1mm}Daria Cherniuk}\\
	Center for Artifical Intelligence Technology\\
	Skolkovo Institute of Science and Technology\\
	Moscow, Russia\\
	\texttt{daria.cherniuk@skoltech.ru} \\
	\And
	\href{https://orcid.org/0000-0002-9274-7237}{\includegraphics[scale=0.06]{orcid.pdf}\hspace{1mm}Aleksandr Mikhalev}\\
	Center for Artifical Intelligence Technology\\
	Skolkovo Institute of Science and Technology\\
	Moscow, Russia\\
	\texttt{al.mikhalev@skoltech.ru} \\
	\And
	\href{https://orcid.org/0000-0003-2071-2163}{\includegraphics[scale=0.06]{orcid.pdf}\hspace{1mm}Ivan Oseledets}\\
    Artificial Intelligence Research Institute \\
    \texttt{oseledets@airi.net} \\
	Center for Artifical Intelligence Technology\\
	Skolkovo Institute of Science and Technology\\
	Moscow, Russia\\
	\texttt{i.oseledets@skoltech.ru} \\
}

\date{}

\renewcommand{\shorttitle}{Run LoRA Run: Faster and Lighter LoRA Implementations}

\hypersetup{
pdftitle={RunLoRA},
pdfsubject={Fast and memory-efficient LoRA adapter},
pdfauthor={Daria Cherniuk, Aleksandr Mikhalev, Ivan Oseledets},
pdfkeywords={},
}

\begin{document}
\maketitle

\begin{abstract}
	LoRA is a technique that reduces the number of trainable parameters in a neural network by introducing low-rank adapters to linear layers. This technique is used both for fine-tuning and full training of large language models. This paper presents the RunLoRA framework for efficient implementations of LoRA that significantly improves the speed of neural network training and fine-tuning using low-rank adapters. The proposed implementation optimizes the computation of LoRA operations based on dimensions of corresponding linear layer, layer input dimensions and lora rank by choosing best forward and backward computation graph based on FLOPs and time estimations, resulting in faster training without sacrificing accuracy. The experimental results show up to 28\% speedup on language modeling networks.
\end{abstract}


\section{Introduction}
\label{sec:intro}
LoRA \cite{Hu2022} paper has introduced the idea of updating a low-rank correction instead of the full weight matrix. This approach quickly became popular due to reduced cost of the update: number of parameters in the adapter is significantly lower than the original due to a low-rank structure.
There emerged a number of papers that proved LoRA's efficacy not only for fine-tuning on downstream tasks, but also for full training (ReLoRA\cite{lialin2023stack}) or style-transfer (ZipLoRA\cite{shah2023ziplora}).
Different modifications of LoRA followed as well, incorporating quantization (QLoRA\cite{Dettmers2023QLoRAEF}), weight-sharing (LoTR\cite{bershatsky2024lotr}, VeRA\cite{kopiczko2024vera}), etc..

However, all variations of LoRA use the default chain of operations while calculating the output, which often leads to sub-optimal graph of computations. And none of the papers on low-rank adapter training considers computation costs. We propose RunLora: a framework which contains different variations of forward and backward pass through an adapter-induced linear layer and chooses the best pair for a given architecture. We provide a thorough analysis (both empirically and theoretically) on areas of applicability of each pass in the parameter space.

Our framework is compatible with pytorch and can be used as a simple model wrapper, similar to Lora implementation from PEFT library. We also provide a functionality to work with quantized model weights to fine-tune models in fasion of QLORA \cite{Dettmers2023QLoRAEF} paper. 

\begin{figure*}[t!]
\includegraphics[width=\textwidth]{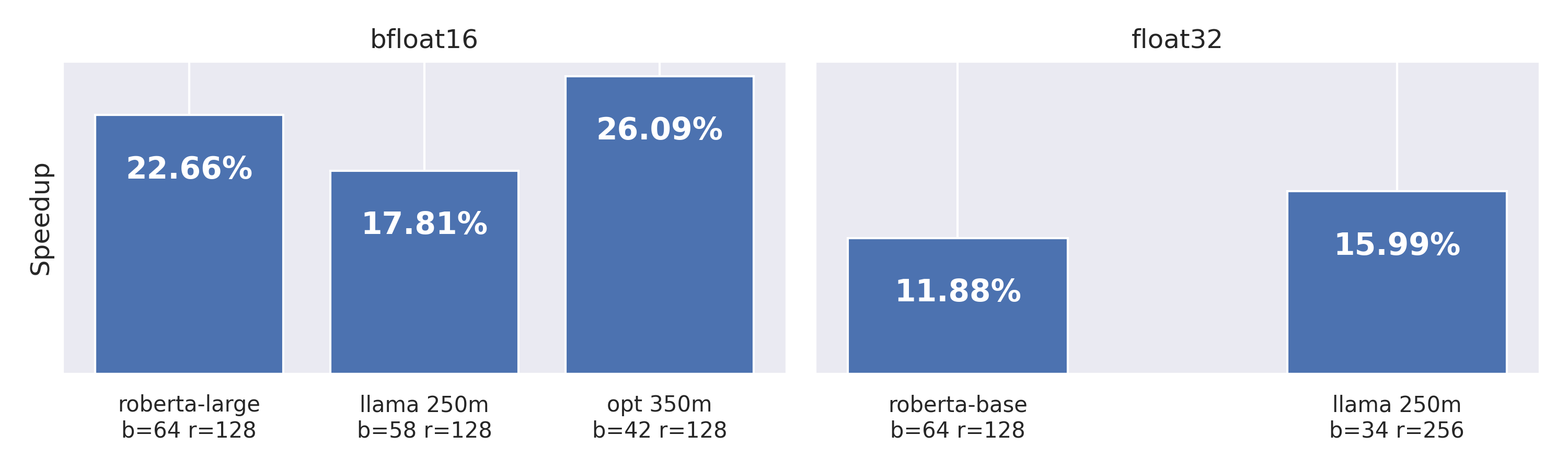}
\centering
\caption{Maximum speedups for forward-backward pass through network achieved on different families of models and with different data types. Note that OPT implementation used in these experiments does not support float32 training. b - batch size, r - lora rank, s - sequence length.}
\label{fig:speedup}
\end{figure*}

We evaluated our framework's performance on a series of NLP models, including RoBerta, OPT and Llama, and achieved up to 28\% speedup (Figure \ref{fig:speedup}) only due to optimized chain of PyTorch operations.
Additionally, we managed to save up to 6Gb of memory due to reduction in number of saved activations (Table \ref{tab:opt_exp}). 

The summary of our contributions is as follows:
\begin{enumerate}
    \item We implemented a number of alternative forward and backward computation passes through low-rank adapters and investigated areas of applicability of each pass under parameter reduction condition. 
    \item We developed a framework called RunLoRA: a model wrapper for training with low-rank adapters which uses the best forward-backward passes considering the model architecture and training parameters. 
    \item We evaluate our framework on a number of common language models: OPT, Llama, RoBerta and demonstrate significant speedups (up to 28\%) proving efficiency of RunLora. 
\end{enumerate}

\section{Problem setting and Methodology}
\label{sec:theory}

Default forward pass of LoRA looks the following:
\begin{equation}
    LoRA(X) = XW + (XA)B,
\end{equation}

The backward pass is then automatically determined by the framework from an autograd technique.
All the optimizations are left to the neural network training framework, which often performs sub-optimally.

Many scientists and engineers avoid the following chain of computations:
\begin{equation}
    LoRA(X) = X(W+AB).
\end{equation}

This is due to an implicit assumption that weights $W$ are quite large and forming a same-size matrix AB is undesirable. 
However, real-world LoRA-adapter training deals with large input $X$ in an attempt of increasing batch size to utilize GPU RAM at its full capacity.
Large batch size leads to a contradiction to the assumption and inefficient LoRA implementation.

Our current implementation contains 2 major and 2 minor variants of the forward pass and 5 major and 30 minor variants of the backward pass.
Major variants are different by opening brackets differently, while minor variants are based on major variants with operations reordered.
Formally, the major forward variants are:
\begin{enumerate}
    \item $Y = (XW)+((XA)B)$
    \item $Y = X(W+AB)$
\end{enumerate}
Unlike default LoRA implementation, neither forward function in RunLoRA saves result of $XA$ to context.


The backward of the LoRA adapter requires us to calculate the following tensors:
\begin{equation}
\left\{ \begin{array}{l}
\mathrm{d} A = X^\top \mathrm{d}Y B^\top,\\
\mathrm{d} B = A^\top X^\top \mathrm{d}Y,\\
\mathrm{d} X = \mathrm{d}Y W^\top + \mathrm{d}Y B^\top A^\top.
\end{array} \right.
\end{equation}
Due to associativity of matrix multiplications, there are several ways of doing it and getting the same result up to rounding errors:
\begin{equation}
    ABC=(AB)C=A(BC).
\end{equation}

There are 3 multiplications and each can be done in 2 ways, which leads to 8 variants of the backward pass:
\begin{enumerate}
    \item $\mathrm{d} A = X^\top (\mathrm{d}Y B^\top)$, \\ $ \mathrm{d} B = (A^\top X^\top) \mathrm{d}Y$, \\$\mathrm{d} X = \mathrm{d}Y W^\top + (\mathrm{d}Y B^\top) A^\top.$
    \item $\mathrm{d} A = X^\top (\mathrm{d}Y B^\top)$, \\$ \mathrm{d} B = A^\top (X^\top \mathrm{d}Y)$, \\$\mathrm{d} X = \mathrm{d}Y W^\top + (\mathrm{d}Y B^\top) A^\top.$
    \item $\mathrm{d} A = (X^\top \mathrm{d}Y) B^\top$, \\$ \mathrm{d} B = A^\top (X^\top \mathrm{d}Y)$, \\$\mathrm{d} X = \mathrm{d}Y W^\top + (\mathrm{d}Y B^\top) A^\top.$
    \item $\mathrm{d} A = (X^\top \mathrm{d}Y) B^\top$, \\$ \mathrm{d} B = A^\top (X^\top \mathrm{d}Y)$, \\$\mathrm{d} X = \mathrm{d}Y (W^\top + B^\top A^\top).$
    \item $\mathrm{d} A = X^\top (\mathrm{d}Y B^\top)$, \\$ \mathrm{d} B = (A^\top X^\top) \mathrm{d}Y $, \\$\mathrm{d} X = \mathrm{d}Y (W^\top + B^\top A^\top).$
    \item $\mathrm{d} A = (X^\top \mathrm{d}Y) B^\top$, \\$ \mathrm{d} B = (A^\top X^\top) \mathrm{d}Y$, \\$\mathrm{d} X = \mathrm{d}Y W^\top + (\mathrm{d}Y B^\top) A^\top.$
    \item $\mathrm{d} A = (X^\top \mathrm{d}Y) B^\top$, \\$ \mathrm{d} B = (A^\top X^\top) \mathrm{d}Y$, \\$\mathrm{d} X = \mathrm{d}Y (W^\top + B^\top A^\top).$
    \item $\mathrm{d} A = X^\top (\mathrm{d}Y B^\top)$, \\$ \mathrm{d} B = A^\top (X^\top \mathrm{d}Y)$, \\$\mathrm{d} X = \mathrm{d}Y (W^\top + B^\top A^\top).$
\end{enumerate}



However, out of eight variants we implement only first five (Algorithms \ref{alg:back1} - \ref{alg:back5}) since others require more or equal number of FLOPs in any setting. Specifically, backward6 would require more FLOPs than backward5 for any architecture and training configuration and backward7, backward8 require the same number of FLOPSs as backward3 (Table \ref{tab:flops}). 

\begin{algorithm}[h!]
\caption{backward1}\label{alg:back1}
\begin{algorithmic}
\State $Z_1 \gets \mathrm{d}YB^\top$
\State $Z_2 \gets XA$
\State $\mathrm{d}A \gets X^\top Z_1$
\State $\mathrm{d}B \gets Z_2^\top \mathrm{d}Y$
\State $\mathrm{d}X \gets \mathrm{d}Y W^\top + Z_1 A^\top$
\end{algorithmic}
\end{algorithm}

\begin{algorithm}[h!]
\caption{backward2}\label{alg:back2}
\begin{algorithmic}
\State $Z_1 \gets \mathrm{d}YB^\top$
\State $Z_2 \gets X^\top \mathrm{d}Y$
\State $\mathrm{d}A \gets X^\top Z_1$
\State $\mathrm{d}B \gets A^\top Z_2$
\State $\mathrm{d}X \gets \mathrm{d}Y W^\top + Z_1 A^\top$
\end{algorithmic}
\end{algorithm}

\begin{algorithm}[h!]
\caption{backward3}\label{alg:back3}
\begin{algorithmic}
\State $Z_1 \gets \mathrm{d}YB^\top$
\State $Z_2 \gets X^\top \mathrm{d}Y$
\State $\mathrm{d}A \gets Z_2 B^\top$
\State $\mathrm{d}B \gets A^\top Z_2$
\State $\mathrm{d}X \gets \mathrm{d}Y W^\top + Z_1 A^\top$
\end{algorithmic}
\end{algorithm}

\begin{algorithm}[h!]
\caption{backward4}\label{alg:back4}
\begin{algorithmic}
\State $Z_1 \gets W+AB$
\State $Z_2 \gets X^\top \mathrm{d}Y$
\State $\mathrm{d}A \gets Z_2 B^\top$
\State $\mathrm{d}B \gets A^\top Z_2$
\State $\mathrm{d}X \gets \mathrm{d}Y Z_1^\top$
\end{algorithmic}
\end{algorithm}

\begin{algorithm}[h!]
\caption{backward5}\label{alg:back5}
\begin{algorithmic}
\State $Z_1 \gets \mathrm{d}YB^\top$
\State $Z_2 \gets XA$
\State $Z_3 \gets W+AB$
\State $\mathrm{d}A \gets X^\top Z_1$
\State $\mathrm{d}B \gets Z_2^\top \mathrm{d}Y$
\State $\mathrm{d}X \gets \mathrm{d}Y Z_3^\top$
\end{algorithmic}
\end{algorithm}

Then we analyse area of applicability for each backward, considering a necessary condition on parameters reduction: number of trainable parameters after LoRA transform should be less than that of the original layer.
\begin{equation}\label{eq:param_red}
    r (i + o) < io
\end{equation}
where $r$ denotes LoRA rank, $i$ and $o$ denote input and output dimensions respectively. 

\begin{figure*}[t!]
     \centering
     \begin{subfigure}[t]{0.47\textwidth}
         \centering
         \includegraphics[width=\textwidth]{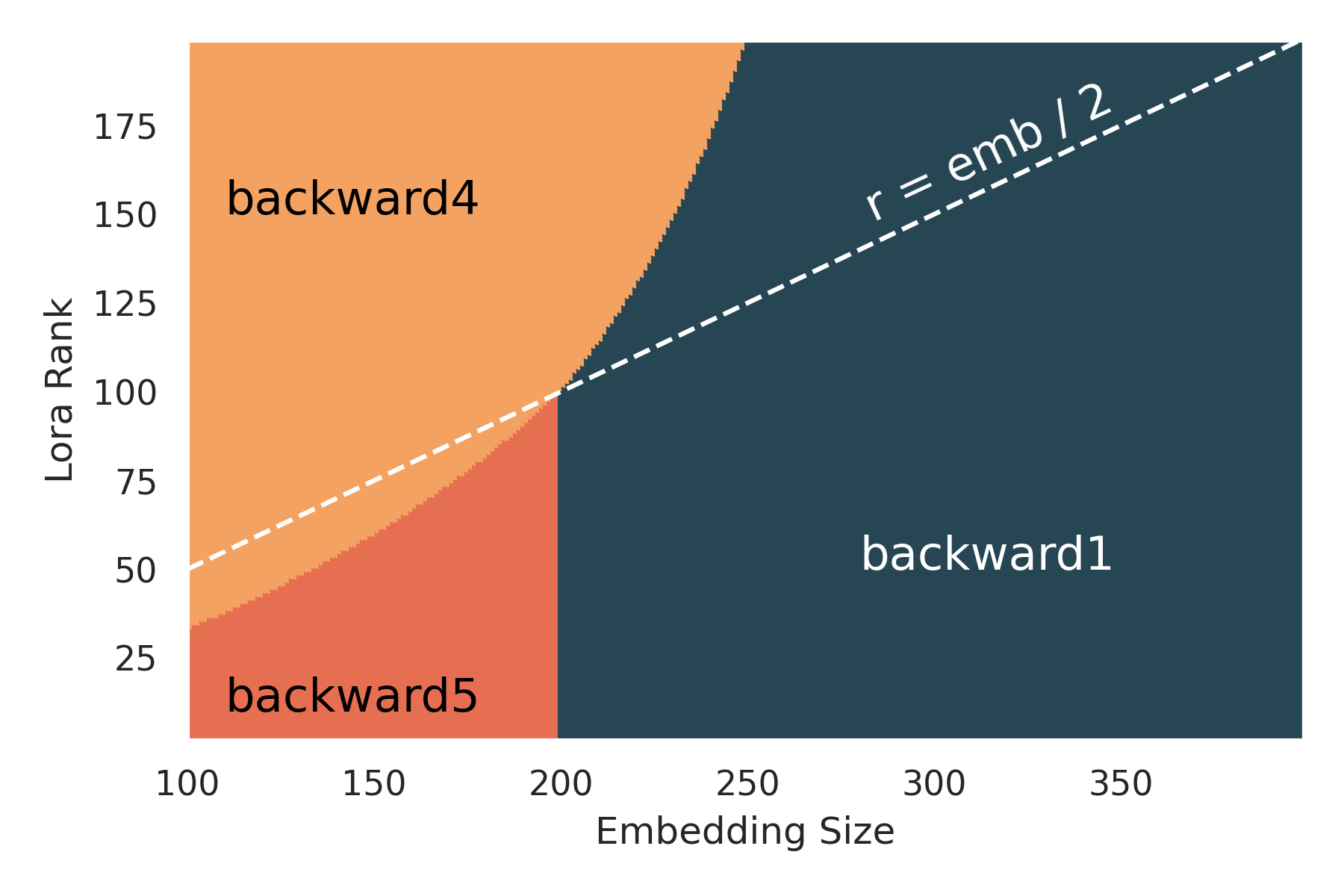}
         \caption{Input and output dimensions are equal to the model's embedding size. batch size = 2, sequence length = 100.}
         \label{fig:areas_ul}
     \end{subfigure}
     \hfill
     \begin{subfigure}[t]{0.47\textwidth}
         \centering
         \includegraphics[width=\textwidth]{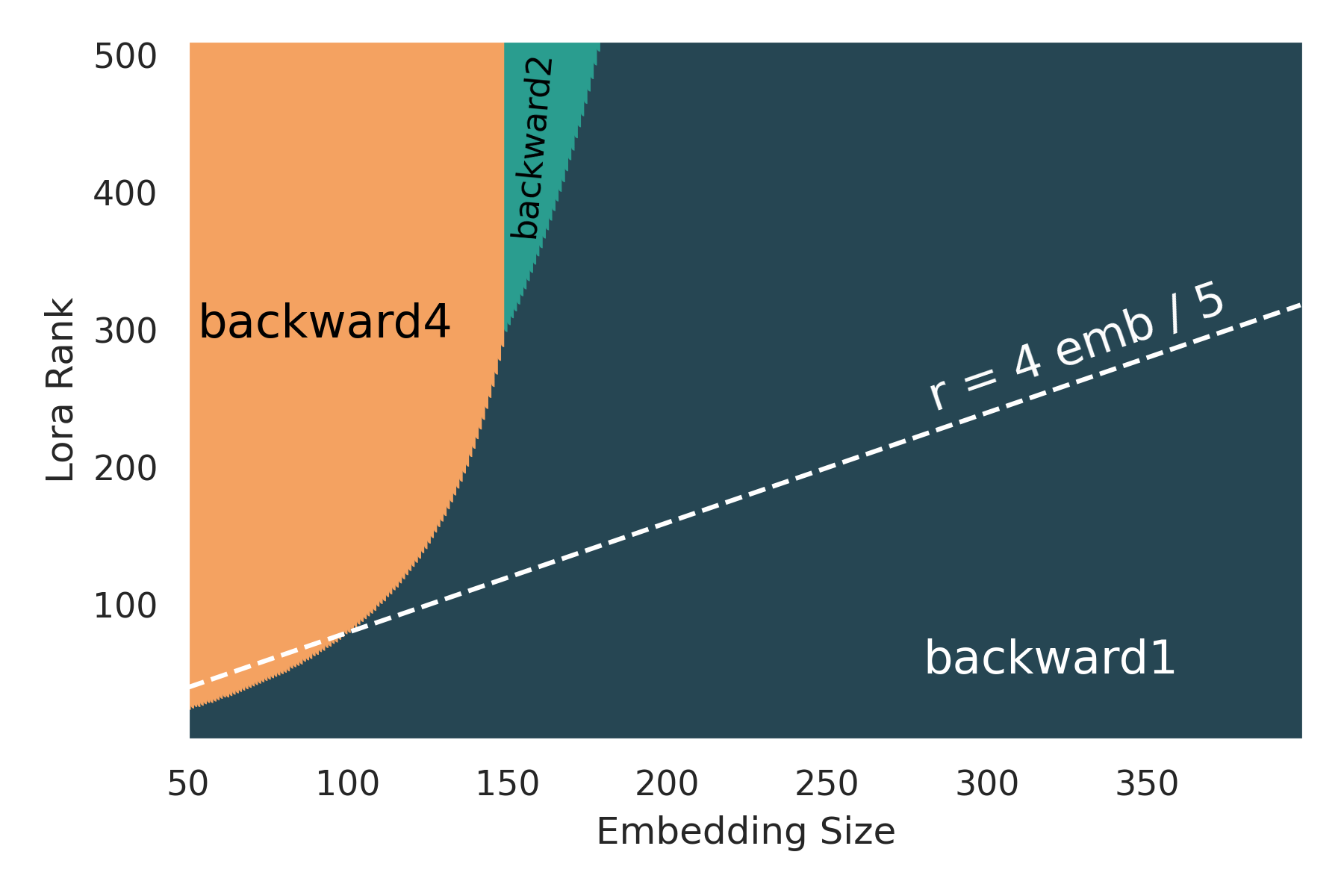}
         \caption{Input dimension equals model's embedding size, output dimension is four times bigger. batch size = 2, sequence length = 100.}
         \label{fig:areas_ur}
     \end{subfigure}
     \hfill
     \par\bigskip
     \begin{subfigure}[t]{0.47\textwidth}
         \centering
         \includegraphics[width=\textwidth]{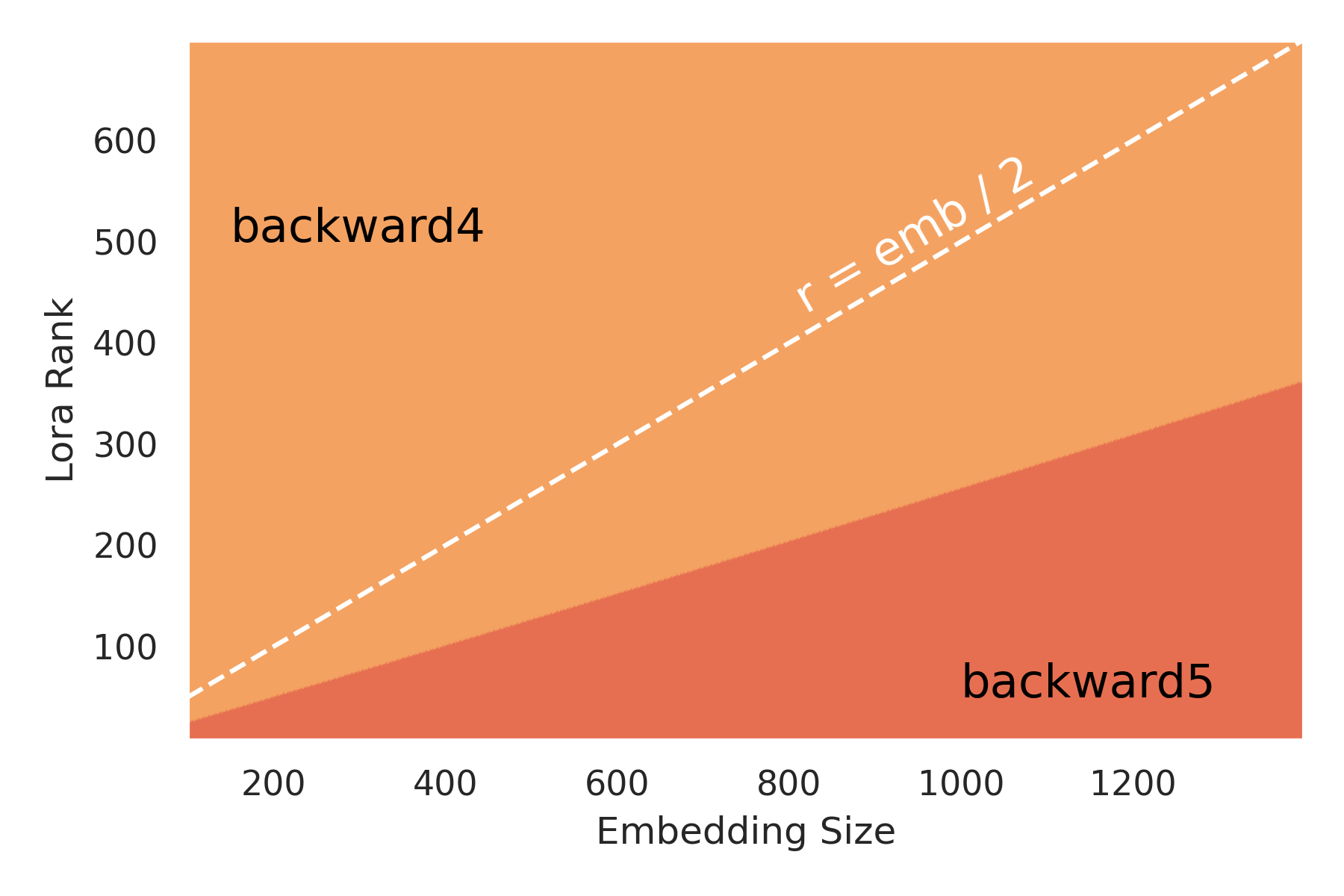}
         \caption{Input and output dimensions are equal to the model's embedding size. batch size = 20, sequence length = 1024.}
         \label{fig:areas_dr}
     \end{subfigure}
     \hfill
     \begin{subfigure}[t]{0.47\textwidth}
         \centering
         \includegraphics[width=\textwidth]{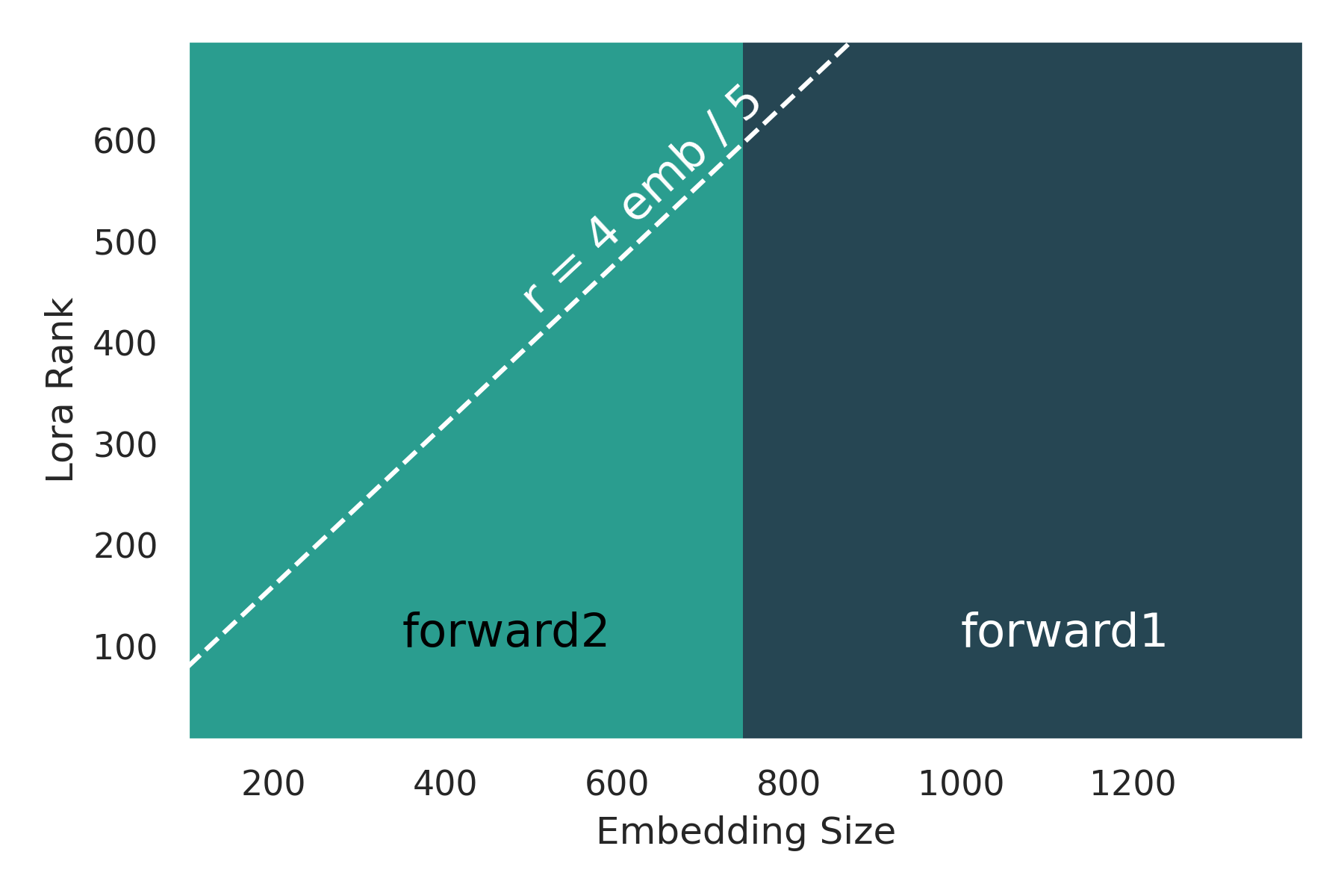}
         \caption{Input dimension equals model's embedding size, output dimension is four times bigger. batch size = 1, sequence length = 600.}
         \label{fig:areas_dl}
     \end{subfigure}
    \par\bigskip
    \caption{Areas of best forward/backward pass choice. Region under the dashed line satisfies condition \ref{eq:param_red}. }
    \label{fig:areas_e_r}
\end{figure*}

Figure \ref{fig:areas_e_r} depicts a case study examples for some batch sizes and sequence lengths. Areas of color depict best choice of forward or backward pass based on minimal number of required FLOPs. Subfigures \ref{fig:areas_ul} and \ref{fig:areas_dl} on the left consider a square weight layer where number of input features and output features equal model's embedding size (i.e., query, key and value layers in transformers). Subfigures \ref{fig:areas_ur} and \ref{fig:areas_dr} on the right depict expanding linear layer (typically, 4x expantion is used in MLP blocks of transformers). 
In all cases, parameter reduction is satisfied only under the dashed line. 

In all depicted cases backward2 and backward3 did not appear as the best choice which satisfies condition \ref{eq:param_red}. It can be further proved that neither backward2 nor backward3 will provide the least number of FLOPs under this restriction. It is sufficient to prove that at least one of other backward algorithms is a better option. For both cases, it is convenient to compare against backward5. 
We will use proof by contradiction:

Suppose FLOPs(backward2) $\leq$ FLOPs(backward5). From \ref{tab:flops} it follows that:
\begin{align*}
    2bs(or + 2ir + 2io) + 2ior &\leq 2bs(2or + 2ir + oi) + 2ior \\
    2bsor + 4bsir + 4bsio + 2ior &\leq 4bsor + 4bsir + 2bsoi + 2ior \\
    2bsio &\leq 2bsor \\
    i &\leq r
\end{align*}

Using \ref{eq:param_red} and knowing that $i > 0, o > 0$:
\begin{align*}
    i \leq r &< \frac{io}{i+o} \le i
\end{align*}
We reached a contradiction. That means FLOPs(backward2) > FLOPs(backward5).

Suppose FLOPs(backward3) $\leq$ FLOPs(backward5). From \ref{tab:flops} it follows that:
\begin{align*}
    2bs(2io + or + ir) +4ior &\leq 2bs(2or + 2ir + oi) + 2ior \\
    4bsio + 2bsor + 2bsir + 4iro &\leq 4bsor + 4bsir + 2bsoi + 2ior \\
    2bsio +2iro &\leq 2bsor + 2bsir \\
    bs(io - or -ir) &\leq -iro
\end{align*}
Using \ref{eq:param_red} and knowing that $i>0, r>0, o > 0, b>0, s>0$:
\begin{align*}
    0 < bs &\leq \frac{-iro}{io - or -ir} < 0
\end{align*}
We reached a contradiction. That means FLOPs(backward3) > FLOPs(backward5).

\begin{table}
    \caption{Number of floating-point operations per second(FLOPs) for our implemented forward and backward passes. $b$ - batch size; $s$ - sequence length; $i$ - input dimension, $o$ - output dimension; $r$ - adapter rank;}
    \centering
    \begin{tabular}{cc}
        \toprule
        Method & FLOPs \\
        \hline \hline
        forward1 & $2b \cdot s \cdot (i \cdot o + r \cdot i + o \cdot r)$\\
        forward2 & $2(i \cdot o \cdot r + b \cdot s \cdot o \cdot i)$\\
        \hline
        backward1 & $2b \cdot s \cdot (2o \cdot r+3i \cdot r+o \cdot i)$\\
        backward2 & $2b \cdot s(o \cdot r+2i \cdot r+2i \cdot o) + 2i \cdot o \cdot r$\\
        backward3 & $2b \cdot s \cdot (2i \cdot o + o \cdot r + i \cdot r) + 4i \cdot r \cdot o$\\
        backward4 & $2 (2b \cdot s \cdot i \cdot o + 3 i \cdot o \cdot r)$\\
        backward5 & $2b \cdot s \cdot (2o \cdot r + 2i \cdot r + o \cdot i) + 2i \cdot o \cdot r$\\
        backward6 & $2b \cdot s \cdot (2o \cdot r + 2i \cdot r + 2o \cdot i)+ 4i \cdot o \cdot r$\\
        backward7 & $2b \cdot s \cdot (o \cdot r + i \cdot r + 2o \cdot i)+ 4i \cdot o \cdot r$\\
        backward8 & $2b \cdot s \cdot (o \cdot r + i \cdot r + 2o \cdot i)+ 4i \cdot o \cdot r$\\
        \bottomrule
    \end{tabular}
    \label{tab:flops}
\end{table}
 
\begin{figure*}[t!]
     \centering
     \begin{subfigure}[t]{0.49\textwidth}
         \centering
         \includegraphics[width=\textwidth]{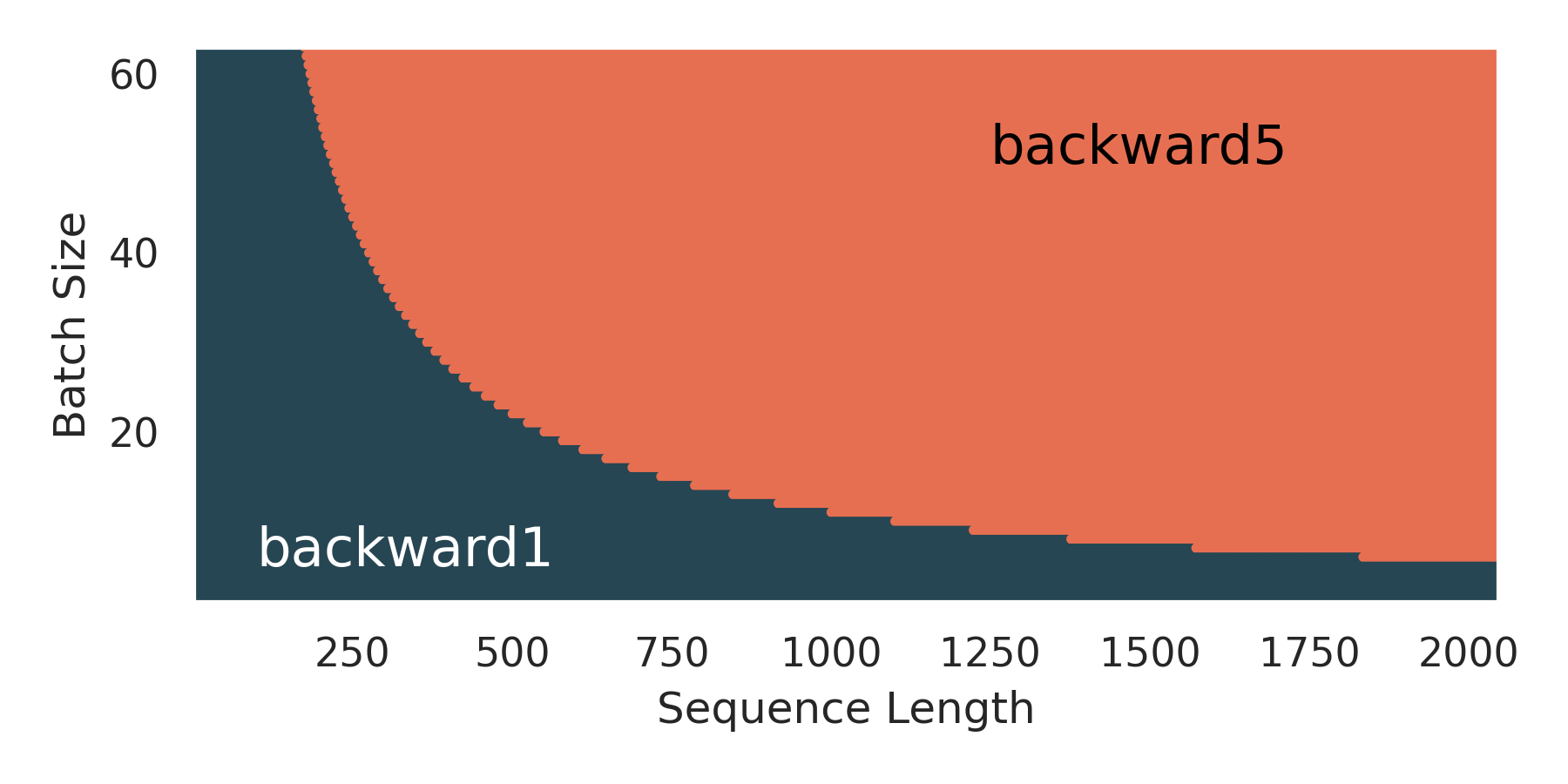}
         \caption{Best backward path as a function of batch size and sequence length.}
         \label{fig:areas_l}
     \end{subfigure}
     \hfill
     \begin{subfigure}[t]{0.49\textwidth}
         \centering
         \includegraphics[width=\textwidth]{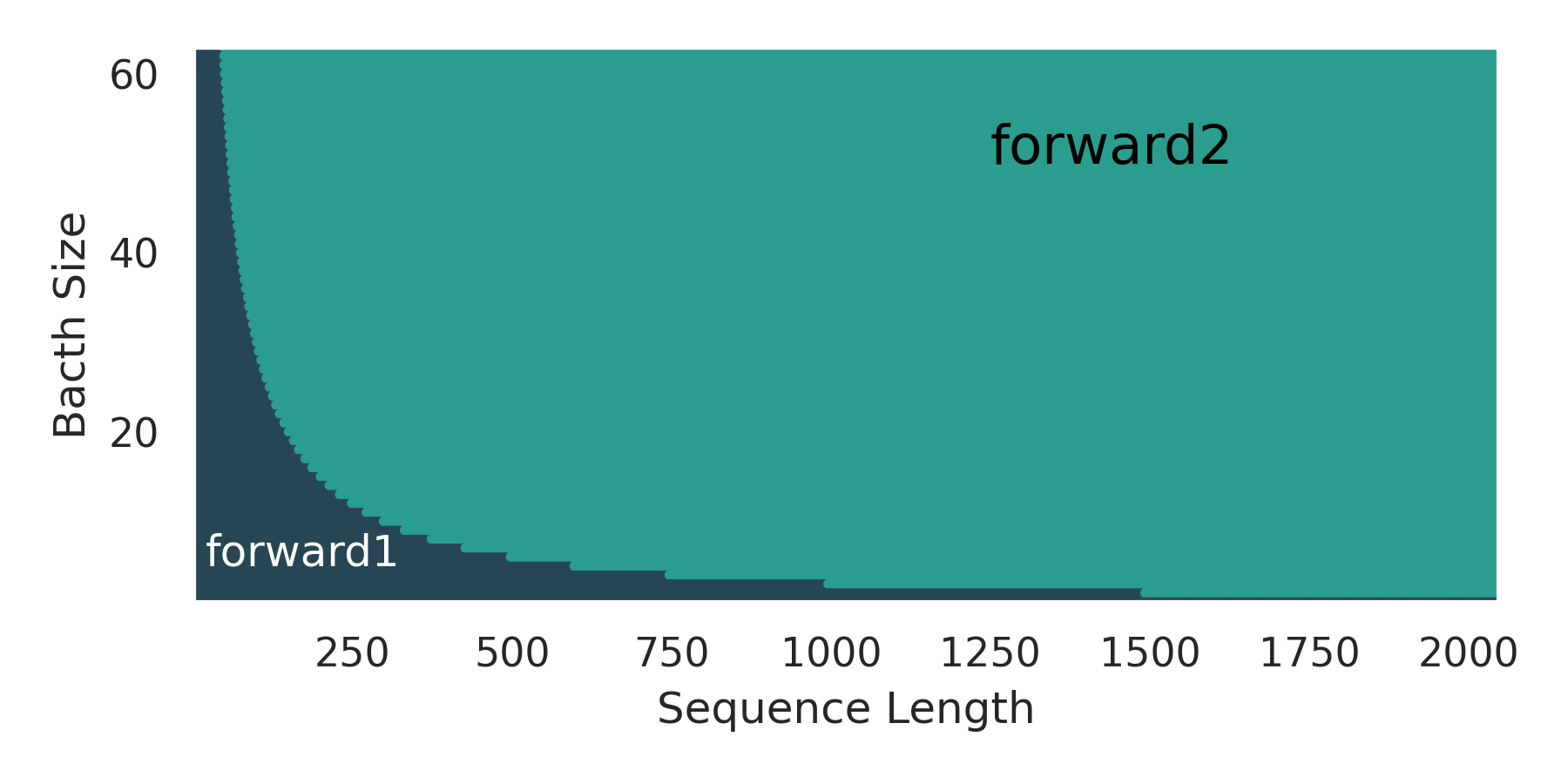}
         \caption{Best forward path as a function of batch size and sequence length.}
         \label{fig:areas_r}
     \end{subfigure}
    \par\bigskip
    \caption{Areas of best forward/backward pass choice for the LlamaMLP linear layer. Input dimention equals 4096, output dimention equals 11008. Rank is 128. }
    \label{fig:areas_b_s}
\end{figure*}

Areas of applicability can also be researched in batch size/sequence length space. For example, Figure \ref{fig:areas_b_s} depicts best backward and forward paths for the LlamaMLP linear layer with adapters of rank 128 (this configuration satisfies condition \ref{eq:param_red}).

\section{Numerical experiments}
\label{sec:experiments}

To evaluate RunLora performance, we have conducted experiments on several NLP Models with number of parameters ranging from 60 millions up to 1.3 billion: LLama~\cite{touvron2023llama}, OPT~\cite{zhang2022opt}, RoBERTa~\cite{liu2020roberta}. 
We measured mean time of a Forward-Backward pass through the network for different architectures and training settings and compared it to Peft LoRA implementation. Additionally, we performed 100 epochs of adapters training on wikitext-2 \cite{wikitext} and compared steps-per-second and samples-per-second metrics as well as total training runtime.  

\paragraph{Llama}
We used Llama model implementation which employs Flash Attention from PyTorch framework. We consider several configurations (from 60 millions of parameters up to 1.3 billions) and initialize layer weights randomly.  
As showed in Table \ref{tab:llama_exp_fp32}, we managed to achieve up to 16\% speedup compared to Peft when running model with float32 data type for weights and operations. 
When running the same experiment in bfloat16 (Table \ref{tab:llama_exp_bf16}) we manage to achieve up to 17.8\% speedup. 
This slight improvement results from the fact that training in bfloat16 is generally faster than training in full precision, which makes reduction of flops due to RunLoRA to be more influencing on the loop runtime. 

When training Llama for 100 epochs on wikitext-2, we manage to achieve 17.56\% reduction in total runtime. Accordingly, number of training samples per second  and number of train steps per seconds increase in 1.2 times (Table ~\ref{tab:llama_training}).

\paragraph{RoBerta} 
Another family of models we consider in our experiments consists of roberta-base and roberta-large pretrained models from Huggingface Hub. They contain about 125 millions and 355 millions of parameters respectively. 
In terms of mean Forward-Backward RunLoRA performs 11.88\% faster in float32 and 22.06\% faster in bloat16 data type. 

As for training RoBerta on wikitext-2, RunLoRA shows up to 20.27\% speedup in total runtime and 1.25 times increase in train samples per second and train steps per second (Table \ref{tab:roberta_training}). 

\paragraph{OPT}
As with RoBerta, we use pretrained weights and model implementation from Huggingface Hub. With OPT models we also use FlashAttention2 \cite{dao2024flashattention} mechanism which only supports bfloat16 data type.  

Table \ref{tab:opt_exp} shows maximum of 28.29\% speedup of Forward-Backward pass for sequence length of 512 and 26.24\% for maximum sequence length of the model. Thereafter, wikitext-2 training experiment (Table \ref{tab:opt_training}) depicts maximum 26.65\% reduction in total runtime, 1.36 increase in train samples per second and train steps per second.

Additionally, since RunLoRA forward functions do not save intermediate result $XA$, in certain experiments we manage to save up to 6Gb of GPU memory.

\paragraph{Llama2-7b}
We also evaluate our framework on a large model with 7 billion parameters. 
Since training such model on a single gpu proves to be a tedious and often impossible task, we use LitGPT\footnote{https://github.com/Lightning-AI/litgpt} framework to get advantages of FSDP training. 
We train Llama2-7b on Alpaca dataset for 100 iteration steps. We use two GPUs with minibatch size of 40. 
Results are presented in Table \ref{tab:llama_7b}: RunLoRA achieves 21.47\% speedup in mean iteration time.

\paragraph{}All experiments were performed on a single Nvidia A100 GPU 80Gb (except for Llama2-7b experiment, which was conducted on two gpus). In all experiments, lora dropout  was fixed to 0, other parameters are stated in the referenced tables. For measuring mean Forward-Backward pass we utilized torch.benchmarking\footnote{https://pytorch.org/docs/stable/benchmark\_utils.html} package. RunLoRA adapters were applied to all linear weights in attention and MLP blocks. 

Note that we use Adam optimizer \cite{adam2014} and torch.autocast\footnote{https://pytorch.org/docs/stable/amp.html} when training adapters. Adam requires memory to store optimizer states and autocast converts some activations to fp32 data type. This forces us to use smaller batch sizes or sequence lengths for wikitext-2 training compared to mean Forward-Backward measurements. 

\begin{table*}[ht!]
    \caption{Comparison between RunLora (with flops criterion) and default LoRA implementation from peft library. $b$-batch size, $r$-lora rank on Llama family of models. Sequence length was set to the maximum sequence length of the model (2048). Experiments were run in float32 data type.}
    \centering
    \scriptsize
    \begin{tabularx}{\textwidth}{X|X|X|X|X}%
    \toprule
    \textbf{Implementation} & \textbf{Mean F-B loop, ms} & \textbf{Memory for F-B loop, MB} & \textbf{Speedup, \%} & \textbf{Memory Saved, MB}\\
    \hline \rowcolor{lightgray} \multicolumn{5}{X}{\textbf{llama 130m, b=64, r=128}}
    \csvreader[head to column names]{llama_130mb64r128.fp32.csv}{}
    {\\\hline \implementation & \meantimeus & \maxmemoverheadMB & \speedup & \memorysavedMB}\\
    \hline \rowcolor{lightgray} \multicolumn{5}{X}{\textbf{llama 250m, b=38, r=128}}
    \csvreader[head to column names]{llama_250mb38r128.fp32.csv}{}
    {\\\hline \implementation & \meantimeus & \maxmemoverheadMB & \speedup & \memorysavedMB}\\
    \hline  \rowcolor{lightgray} \multicolumn{5}{X}{\textbf{llama 250m, b=34, r=256}}
    \csvreader[head to column names]{llama_250mb34r256.fp32.csv}{}
    {\\\hline \implementation & \meantimeus & \maxmemoverheadMB & \speedup & \memorysavedMB}\\
    \hline  \rowcolor{lightgray}  \multicolumn{5}{X}{\textbf{llama 350m, b=30, r=256}}
    \csvreader[head to column names]{llama_350mb30r256.fp32.csv}{}
    {\\\hline \implementation & \meantimeus & \maxmemoverheadMB & \speedup & \memorysavedMB}\\
    \end{tabularx}
    \label{tab:llama_exp_fp32}
\end{table*}

\begin{table*}[ht!]
    \caption{Comparison between RunLora (with flops criterion) and default LoRA implementation from peft library. $b$-batch size, $r$-lora rank on Llama family of models. Sequence length was set to the maximum sequence length of the model (2048). Experiments were run in bfloat16 data type.}
    \centering
    \scriptsize
    \begin{tabularx}{\textwidth}{X|X|X|X|X}%
    \toprule
    \textbf{Implementation} & \textbf{Mean F-B loop, ms} & \textbf{Memory for F-B loop, MB} & \textbf{Speedup, \%} & \textbf{Memory Saved, MB}\\
    \hline \rowcolor{lightgray}  \multicolumn{5}{X}{\textbf{llama 130m, b=64, r=8}}
    \csvreader[head to column names]{llama_130mb64r8.csv}{}
    {\\\hline \implementation & \meantimeus & \maxmemoverheadMB & \speedup & \memorysavedMB}\\
    \hline \rowcolor{lightgray}  \multicolumn{5}{X}{\textbf{llama 250m, b=58, r=128}}
    \csvreader[head to column names]{llama_250mb58r128.csv}{}
    {\\\hline \implementation & \meantimeus & \maxmemoverheadMB & \speedup & \memorysavedMB}\\
    \hline \rowcolor{lightgray}  \multicolumn{5}{X}{\textbf{llama 350m, b=48, r=128}}
    \csvreader[head to column names]{llama_350mb48r128.csv}{}
    {\\\hline \implementation & \meantimeus & \maxmemoverheadMB & \speedup & \memorysavedMB}\\
    \hline \rowcolor{lightgray}  \multicolumn{5}{X}{\textbf{llama 1.3b, b=24, r=512}}
    \csvreader[head to column names]{llama_1bb24r512.csv}{}
    {\\\hline \implementation & \meantimeus & \maxmemoverheadMB & \speedup & \memorysavedMB}
    \end{tabularx}
    \label{tab:llama_exp_bf16}
\end{table*}

\begin{table*}[ht!]
    \caption{Comparison between RunLora (with flops criterion) and default LoRA implementation from peft library. $b$-batch size, $r$-lora rank on OPT family of models. Sequence length was set to the maximum sequence length of the model (2048).  Experiments were run in bfloat16
data type.}
    \centering
    \scriptsize
    \begin{tabularx}{\textwidth}{X|X|X|X|X}%
    \toprule
    \textbf{Implementation} & \textbf{Mean F-B loop, ms} & \textbf{Memory for F-B loop, MB} & \textbf{Speedup, \%} & \textbf{Memory Saved, MB}\\
    \hline  \rowcolor{lightgray} \multicolumn{5}{X}{\textbf{opt 125m, b=64, r=8}}
    \csvreader[head to column names]{opt-125mb64r8.csv}{}
    {\\\hline \implementation & \meantimeus & \maxmemoverheadMB & \speedup & \memorysavedMB}\\
    \hline  \rowcolor{lightgray} \multicolumn{5}{X}{\textbf{opt 350m, b=42, r=128}}
    \csvreader[head to column names]{opt-350mb42r128.csv}{}
    {\\\hline \implementation & \meantimeus & \maxmemoverheadMB & \speedup & \memorysavedMB}\\
    \hline  \rowcolor{lightgray} \multicolumn{5}{X}{\textbf{opt 1.3b, b=24, r=128}}
    \csvreader[head to column names]{opt-1.3bb24r128.csv}{}
    {\\\hline \implementation & \meantimeus & \maxmemoverheadMB & \speedup & \memorysavedMB}\\
    \hline \rowcolor{lightgray}  \multicolumn{5}{X}{\textbf{opt 1.3b, b=22, r=512}}
    \csvreader[head to column names]{opt-1.3bb22r512.csv}{}
    {\\\hline \implementation & \meantimeus & \maxmemoverheadMB & \speedup & \memorysavedMB}\\
    \end{tabularx}
    \label{tab:opt_exp}
\end{table*}

\begin{table*}[ht!]
    \caption{Comparison between RunLora (with flops criterion) and default LoRA implementation from peft library. $b$-batch size, $r$-lora rank on OPT family of models. Sequence length was set to 512. Experiments were run in bfloat16
data type.}
    \centering
    \scriptsize
    \begin{tabularx}{\textwidth}{X|X|X|X|X}%
    \toprule
    \textbf{Implementation} & \textbf{Mean F-B loop, ms} & \textbf{Memory for F-B loop, MB} & \textbf{Speedup, \%} & \textbf{Memory Saved, MB}\\
\hline  \rowcolor{lightgray} \multicolumn{5}{X}{\textbf{opt-125m, b=64, r=128}}
\csvreader[head to column names]{opt-125mb64s512r128.bf16.csv}{}
{\\\hline \implementation & \meantimeus & \maxmemoverheadMB & \speedup & \memorysavedMB}\\
\hline  \rowcolor{lightgray} \multicolumn{5}{X}{\textbf{opt-350m, b=64, r=128}}
\csvreader[head to column names]{opt-350mb64s512r128.bf16.csv}{}
{\\\hline \implementation & \meantimeus & \maxmemoverheadMB & \speedup & \memorysavedMB}\\
\hline  \rowcolor{lightgray} \multicolumn{5}{X}{\textbf{opt-350m, b=100, r=128}}
\csvreader[head to column names]{opt-350mb100s512r128.bf16.csv}{}
{\\\hline \implementation & \meantimeus & \maxmemoverheadMB & \speedup & \memorysavedMB}\\
\hline  \rowcolor{lightgray} \multicolumn{5}{X}{\textbf{opt-1.3, b=64, r=128}}
\csvreader[head to column names]{opt-1.3bb64s512r128.bf16.csv}{}
{\\\hline \implementation & \meantimeus & \maxmemoverheadMB & \speedup & \memorysavedMB}\\
\hline  \rowcolor{lightgray} \multicolumn{5}{X}{\textbf{opt-1.3, b=100, r=128}}
\csvreader[head to column names]{opt-1.3bb100s512r128.bf16.csv}{}
{\\\hline \implementation & \meantimeus & \maxmemoverheadMB & \speedup & \memorysavedMB}\\
    \end{tabularx}
    \label{tab:opt_exp_s512}
\end{table*}

\begin{table*}[ht!]
    \caption{Comparison between RunLora (with flops criterion) and default LoRA implementation from peft library. $b$-batch size, $r$-lora rank on RoBERTa family of models. Sequence length was set to the maximum sequence length of the model(512). Experiments were run in float32 data type.}
    \centering
    \scriptsize
    \begin{tabularx}{1.01\textwidth}{X|X|X|X|X}%
    \toprule
    \textbf{Implementation} & \textbf{Mean F-B loop, ms} & \textbf{Memory for F-B loop, MB} & \textbf{Speedup, \%} & \textbf{Memory Saved, MB}\\
\hline  \rowcolor{lightgray} \multicolumn{5}{X}{\textbf{roberta-base, b=64, r=128}}
\csvreader[head to column names]{roberta-baseb64r128.fp32.csv}{}
{\\\hline \implementation & \meantimeus & \maxmemoverheadMB & \speedup & \memorysavedMB}\\
\hline  \rowcolor{lightgray} \multicolumn{5}{X}{\textbf{roberta-large, b=50, r=128}}
\csvreader[head to column names]{roberta-largeb50r128.fp32.csv}{}
    {\\\hline \implementation & \meantimeus & \maxmemoverheadMB & \speedup & \memorysavedMB}\\
    \end{tabularx}
    \label{tab:roberta_exp_fp32}
\end{table*}

\begin{table*}[ht!]
    \caption{Comparison between RunLora (with flops criterion) and default LoRA implementation from peft library. $b$-batch size, $r$-lora rank on RoBERTa family of models. Sequence length was set to the maximum sequence length of the model(512). Experiments were run in bfloat16 data type.}
    \centering
    \scriptsize
    \begin{tabularx}{1.01\textwidth}{X|X|X|X|X}%
    \toprule
    \textbf{Implementation} & \textbf{Mean F-B loop, ms} & \textbf{Memory for F-B loop, MB} & \textbf{Speedup, \%} & \textbf{Memory Saved, MB}\\
\hline  \rowcolor{lightgray} \multicolumn{5}{X}{\textbf{roberta-base, b=64, r=128}}
\csvreader[head to column names]{roberta-baseb64r128.bf16.csv}{}
{\\\hline \implementation & \meantimeus & \maxmemoverheadMB & \speedup & \memorysavedMB}\\
\hline  \rowcolor{lightgray} \multicolumn{5}{X}{\textbf{roberta-large, b=64, r=128}}
\csvreader[head to column names]{roberta-largeb64r128.bf16.csv}{}
{\\\hline \implementation & \meantimeus & \maxmemoverheadMB & \speedup & \memorysavedMB}\\
    \end{tabularx}
    \label{tab:roberta_exp_bf16}
\end{table*}

\begin{table*}[ht!]
    \caption{Training Llama Causal LM model for 100 epochs on wikitext-2 dataset. Training was conducted in bfloat16 data type on a single A100 GPU. Texts are grouped to batches with sequence length of 1024 tokens. }
    \centering
    \scriptsize
    \begin{tabularx}{\textwidth}{X|X|X|X|X}%
    \toprule
    \textbf{Implementation} & \textbf{Train Samples per Second} & \textbf{Train Steps per Second}  & \textbf{Train Runtime, Min} & \textbf{Speedup, \%} \\
    \hline \rowcolor{lightgray} \multicolumn{5}{X}{\textbf{llama-130m, b=60, r=128}} \\
    \hline PEFT & 89.46 & 1.51 & 51.91 & - \\
    \hline RunLoRA & 103.9 & 1.75 & 44.69 & 13.91 \\
    \hline \rowcolor{lightgray} \multicolumn{5}{X}{\textbf{llama-250m, b=48, r=128}} \\
    \hline PEFT & 47.1 & 0.99 & 98.54 & - \\
    \hline RunLoRA & 56.2 & 1.19 & 82.62 & 16.16 \\
    \hline \rowcolor{lightgray} \multicolumn{5}{X}{\textbf{llama-350m, b=40, r=128}} \\
    \hline PEFT & 38.1 & 0.96 & 121.98 & - \\
    \hline RunLoRA & 46.2 & 1.16 & 100.56 & 17.56 \\
    \hline \rowcolor{lightgray} \multicolumn{5}{X}{\textbf{llama-1.3b, b=24, r=128}} \\
    \hline PEFT & 11.18 & 0.47 & 415.25 & - \\
    \hline RunLoRA & 12.46 & 0.52 & 372.69 & 10.2 \\
    \end{tabularx}
    \label{tab:llama_training}
\end{table*}

\begin{table*}[ht!]
    \caption{Training RoBerta Causal LM model for 100 epochs on wikitext-2 dataset. Training was conducted in bfloat16 data type on a single A100 GPU. Texts are grouped to batches with sequence length of 512 tokens. }
    \centering
    \scriptsize
    \begin{tabularx}{1.01\textwidth}{X|X|X|X|X}%
    \toprule
    \textbf{Implementation} & \textbf{Train Samples per Second} & \textbf{Train Steps per Second} & \textbf{Train Runtime, Min} & \textbf{Speedup, \%} \\
    \hline \rowcolor{lightgray} \multicolumn{5}{X}{\textbf{roberta-base, b=64, r=128}} \\
    \hline PEFT & 95.84 & 1.5 & 85.4 & - \\
    \hline RunLoRA & 115.04 & 1.8 & 71.17 & 16.7 \\
    \hline \rowcolor{lightgray} \multicolumn{5}{X}{\textbf{roberta-large, b=46, r=128}} \\
    \hline PEFT &  42.79 & 0.93 & 186.87 & - \\
    \hline RunLoRA & 53.67 & 1.18 & 148.99 & 20.27 \\
    \end{tabularx}
    \label{tab:roberta_training}
\end{table*}

\begin{table*}[ht!]
    \caption{Training OPT Causal LM model for 100 epochs on wikitext-2 dataset. Training was conducted in bfloat16 data type on a single A100 GPU. Texts are grouped to batches with sequence length of 1024 tokens. }
    \centering
    \scriptsize
    \begin{tabularx}{\textwidth}{X|X|X|X|X}%
    \toprule
    \textbf{Implementation} & \textbf{Train Samples per Second} & \textbf{Train Steps per Second} & \textbf{Train Runtime, Min} & \textbf{Speedup, \%} \\
    \hline \rowcolor{lightgray} \multicolumn{5}{X}{\textbf{opt-125m, b=48, r=128}} \\
    \hline PEFT & 75.63 & 1.6 & 51.89 & - \\
    \hline RunLoRA & 96.85 & 2.1 & 40.53 & 21.89 \\
    \hline \rowcolor{lightgray} \multicolumn{5}{X}{\textbf{opt-350m, b=32, r=128}} \\
    \hline PEFT &  34.07 & 1.07 & 115.19 & - \\
    \hline RunLoRA & 46.45 & 1.46 & 84.49 & 26.65 \\
    \hline \rowcolor{lightgray} \multicolumn{5}{X}{\textbf{opt-1.3b, b=20, r=128}} \\
    \hline PEFT &  15.81 & 0.79 & 248.29 & - \\
    \hline RunLoRA & 20.03 & 1.0 & 196.01 & 21.05 \\
    \end{tabularx}
    \label{tab:opt_training}
\end{table*}

\begin{table*}[ht!]
    \caption{Training Llama2-7b model for 100 iterations on Alpaca dataset. Training was conducted in bfloat16 data type on two A100 GPUs. Texts were grouped to batches with sequence length of 512 tokens. Minibatch size = 40. LoRA rank = 128. }
    \centering
    \scriptsize
    \begin{tabularx}{\textwidth}{X|X|X|X|X}%
    \toprule
    \textbf{Implementation} & \textbf{Mean Iteration Time, Sec} & \textbf{Speedup, \%} & \textbf{Train Runtime, Sec} & \textbf{Memory used, GB} \\
    \hline PEFT & 7283.90 & - & 719.76 & 24.09 \\
    \hline RunLoRA & 5720.12 & 21.47 & 573.35 & 23.72 \\
    \end{tabularx}
    \label{tab:llama_7b}
\end{table*}

\section{Conclusion and Future Work}
\label{sec:conclusion}

We have proposed several computation algorithms alternative to a default forward-backward pass through low-rank adapters and derived theoretical bounds of their applicability. 
We have implemented the proposed methods as pytorch-compatible framework RunLoRA which chooses the best computation graph based on model architecture and training parameters. 
We have proved RunLoRA's efficiency by comparing to the peft LoRA implementation.

One of the possible directions for future work is finding optimal computation graphs for approximate versions of low-rank adapters (for example, vector analogues like VeRA\cite{kopiczko2024vera}, DoRA\cite{liu2024dora}).

\section{Related Work}
\label{sec:related_work}

Introduction of LoRA \cite{Hu2022} caused a wave of new publications on topic of low-rank updates. 
For example, ReLoRA\cite{lialin2023stack} has devised a special learning rate scheduler for full training with low-rank updates, ZipLoRA\cite{shah2023ziplora} merges adapters trained separately for style and object, performing effective style transfer, DyLoRA\cite{valipour-etal-2023-dylora} trains LoRA blocks for a range of ranks instead of a single rank.

Many papers try to further reduce the costs of training.
QLoRA\cite{Dettmers2023QLoRAEF} utilizes adapters together with quantization of original weights to reduce memory requirements, 
Vector-based Random Matrix Adaptation (VeRA) \cite{kopiczko2024vera} reduces number of trainable parameters by using a single pair of low-rank matrices shared across all layers and learning small scaling vectors instead. 
LoTR \cite{bershatsky2024lotr} also proposes weight sharing for factors in Tucker2 decomposition of low-rank adapters.
LoRA-FA\cite{zhang2023lorafa} aimes to reduce memory consumption by freezing downscaling half of the LoRA adapters. 

Our method also tries to further increase the efficiency of low-rank adapters training, but with different approach: we nigher reduce the number of LoRA parameters nor compromise training accuracy. Our framework achieves computational speedups and memory reduction only due to the choice of optimal computation graph.

\bibliographystyle{unsrtnat}
\bibliography{main}

\end{document}